%% file: Arxiv2027.tex
\documentclass[letterpaper]{article}
\usepackage[preprint]{aaai2027}
\usepackage[hyphens]{url}
\usepackage{graphicx}
\usepackage{natbib}
\usepackage{caption}
\usepackage{amsmath}
\usepackage{amssymb}
\usepackage{booktabs}
\usepackage{algorithm}
\usepackage{algorithmic}
\newcommand{\method}{BagShift}

\title{BagShift: Measuring How Patch Selection Changes the Evidence Seen by Whole-Slide MIL}
\author{
Ruicheng Yuan\textsuperscript{\rm 1,\rm 2},
Zhenxuan Zhang\textsuperscript{\rm 2},
Liwei Hu\textsuperscript{\rm 2},
Anbang Wang\textsuperscript{\rm 2},\\
Haijie Xu\textsuperscript{\rm 3},
Jiawei Luo\textsuperscript{\rm 1}\corresponding,
Guang Yang\textsuperscript{\rm 2}\corresponding
}
\affiliations{
\textsuperscript{\rm 1}College of Computer Science and Electronic Engineering, Hunan University, Changsha, Hunan, China\\
\textsuperscript{\rm 2}Department of Bioengineering and Imperial-X, Imperial College London, London, UK\\
raytion@hnu.edu.cn, luojiawei@hnu.edu.cn, g.yang@imperial.ac.uk\\
}

\begin{document}
\maketitle

\begin{abstract}
Whole-slide multiple-instance learning (MIL) observes only the patches admitted by its selector. Deployment can alter this selector through compute limits, tissue masking, or regional workflows, even when the patch count is unchanged. We introduce \method{}, a paired protocol that changes the selector for the same case while holding its features and predictor fixed, thereby isolating selector response from case mix. With equal 128-patch budgets, sampling across the tissue or concentrating around one coordinate exposes markedly different evidence: on PANDA, the two views reduce quadratic weighted kappa by 1.57 and 17.96 points, respectively (QWK reported on the $\times100$ scale). On CAMELYON16, lesion annotations withheld from model development show that localized views retain tumor in only 10.0\% of micrometastatic observations, and matched exposure does not consistently recover the loss. The same fixed-count stressor produces a much smaller response on external lung subtyping, although differences in relative coverage make cross-task severity descriptive. When repeated localized observations are available, unioning their patches before one nonlinear MIL pass improves PANDA QWK by 7.87 points over averaging regional predictions. Patch count specifies computation, not observed evidence; deployment evaluations should report both what a selector preserves and how repeated observations are aggregated.
\end{abstract}

\section{Introduction}

Weakly supervised computational pathology predicts from a sample of a whole-slide image (WSI), not from the slide itself. Attention MIL~\cite{ilse2018attention}, large weakly supervised systems~\cite{campanella2019clinical}, and aggregators designed for pathology~\cite{lu2021clam,li2021dsmil,shao2021transmil} learn diagnoses at slide level from bags of patch embeddings. Foundation encoders improve the representation of each patch~\cite{chen2024uni,xu2024gigapath,vorontsov2024virchow}, but a separate selection rule still determines which tissue reaches the predictor.

Deployment can change that rule. A compute budget may thin patches across the slide, while a tissue mask or regional workflow may remove entire areas. These changes are often summarized by the number of retained patches. Count, however, says nothing about where those patches lie: 128 dispersed patches can span the tissue, whereas 128 neighboring patches can fit inside one region. The bags cost the same to process but may contain different diagnostic evidence.

The claim is not merely that broader coverage can help. Patch cardinality specifies computational cost, but not spatial coverage or the observation channel placed in front of MIL. It therefore cannot predict how the model will respond when that channel changes.

Existing evaluations leave this effect difficult to identify. Cohort shift benchmarks change the cases while holding bag construction fixed \cite{koh2021wilds}; methods for selective examination learn a preferred selector \cite{tang2023mhim,neidlinger2026eagle,zhang2026truecam}. A deployment analysis needs the complementary comparison: change the selector for the same slide and measure how the fixed predictor responds. This comparison separates selection from case mix and turns patch construction into an auditable part of the deployed model.

\begin{figure*}[!t]
\centering
\includegraphics[width=\textwidth]{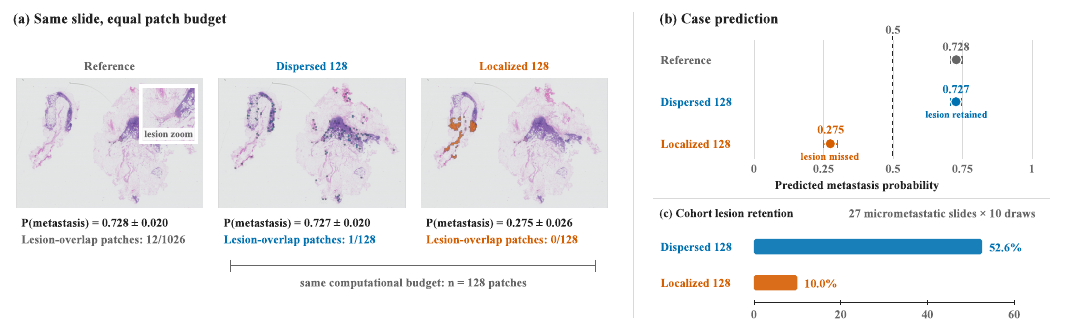}
\caption{Equal patch budgets can retain different focal evidence. (a) The same CAMELYON16 micrometastatic slide under reference, dispersed-128, and localized-128 selection; magenta marks annotated tumor and colored overlays mark retained patches. Both restricted bags contain 128 patches but retain one versus zero tumor-overlap patches. (b) Metastasis probability across five model seeds. (c) Tumor-retention frequency across 27 micrometastatic slides and ten selector draws. The deterministic case-selection rule is given in the supplement.}
\label{fig:overview}
\end{figure*}

We introduce \method{} for this purpose. Each policy acts on the same slide, label, frozen features, and predictor; only the observed patch set changes. The resulting paired response curve reveals whether a model is sensitive to random thinning, a fixed budget, or localized coverage. Matched training controls then ask how much of that response can be reduced through exposure or loss weighting. When several localized observations are available, the same framework compares two inference rules: average their predictions, or unite their patches before one nonlinear MIL pass. Figure~\ref{fig:overview} shows this failure chain on one slide whose lesion annotation was withheld from model development: bags of equal size retain different focal evidence, produce different predictions, and reproduce the same retention ordering at the cohort level.

The experiments follow the resulting evidence chain. PANDA establishes the geometry gap at equal counts. CAMELYON16 uses lesion annotations withheld from model development to test whether the larger response coincides with lost focal evidence, whereas TCGA-to-CPTAC supplies a boundary case with a small response. A multi-image study at two hospitals provides a scope check for semantic selection. Finally, experiments with repeated views show that recovering coverage is insufficient once regional predictions have been pooled separately.

Our contributions are:
\begin{itemize}
\item We formulate selection-policy shift as an intervention within each case and introduce paired policy response curves that isolate it from case mix.
\item Across three public WSI tasks, we show that equal patch counts can preserve unequal evidence. CAMELYON16 annotations withheld from model development connect the largest response to missed micrometastatic tissue and reveal a failure case for matched exposure under inherited slide labels.
\item We identify a practical consequence for repeated observations: union available patches before nonlinear MIL aggregation rather than averaging predictions from separate regions.
\end{itemize}

\section{Related Work}

\paragraph{Whole-slide MIL.}
Attention pooling provides a trainable permutation-invariant aggregator \cite{ilse2018attention}. CLAM adds instance clustering~\cite{lu2021clam}, DSMIL combines critical-instance and bag streams~\cite{li2021dsmil}, and TransMIL models correlated instances with transformer layers and PPEG \cite{shao2021transmil}. DTFD-MIL creates pseudo-bags \cite{zhang2022dtfd}, interventional MIL targets contextual confounding \cite{lin2023interventional}, and MHIM-MIL masks hard instances during teacher--student training~\cite{tang2023mhim}. These methods change the aggregator or training instances. BagShift holds the diagnostic target and encoder fixed and changes the policy that exposes instances.

\paragraph{Selection and robustness.}
Patch-selection methods ask which instances a model should retain. SI-MIL couples interpretable and deep features~\cite{kapse2024simil}; TRUECAM removes ambiguous tiles~\cite{zhang2026truecam}; EAGLE examines compact informative regions~\cite{neidlinger2026eagle}; and PAMIL, EvoPS, and AdaSlide learn efficient or spatially diverse subsets \cite{zheng2024pamil,hashemian2025evops,lee2025adaslide}. Random-sampling and dropout studies vary how many instances survive training or inference \cite{keshvarikhojasteh2024random,zhu2025mildropout}. These approaches seek a useful sampling regime. BagShift asks a different deployment question: how does a trained predictor respond when an external selector changes the observed bag for the same case? The paired intervention distinguishes this response from hospital or scanner shift, which changes the case or acquisition distribution. Group DRO and CVaR motivate two diagnostic training controls \cite{sagawa2020groupdro,rockafellar2000cvar}; the measurement protocol itself does not require a robust optimization objective.

\section{Selection-Policy Shift}

Let $S_i$ denote an underlying slide. A reference extraction channel $A_0$ produces a bag of frozen embeddings and coordinates, while an observation policy $p$ exposes a nonempty subbag. For selector draw $\omega$,
\begin{equation}
 \begin{aligned}
 B_i&=A_0(S_i)=\{(\mathbf{x}_{ij},\mathbf{c}_{ij})\}_{j=1}^{n_i},\\
 B_i^{p,\omega}
 &=\{(\mathbf{x}_{ij},\mathbf{c}_{ij}):j\in I_p(B_i;\omega)\}
 \sim q_p(\cdot\mid B_i),
 \end{aligned}
 \label{eq:observation-channel}
\end{equation}
where $\mathbf{x}_{ij}\in\mathbb{R}^d$, $\mathbf{c}_{ij}$ is its coordinate, and $\varnothing\ne I_p(B_i;\omega)\subseteq[n_i]$. We suppress $\omega$ when the particular draw is not material. Equation~\ref{eq:observation-channel} isolates the selector as the intervention: the slide, label, and shared predictor remain fixed. Here a \emph{policy} is a selection distribution, a \emph{selector draw} is one realization, and the resulting subbag is the observed \emph{view}. A policy may be exogenous and random, coordinate-dependent, or semantic, as in an ROI selected by a pathologist. The experiments use controlled exogenous and coordinate-dependent policies; model-dependent attention removal is only a supplement diagnostic.

For example, with budget $b$ and anchor $a\sim\mathrm{Unif}([n_i])$, the spatial policy is
\begin{equation}
I_{\mathrm{sp}}(B_i;\omega) =\underset{I\subseteq[n_i],\ |I|=\min(b,n_i)}{\arg\min} \sum_{j\in I}\|\mathbf{c}_{ij}-\mathbf{c}_{ia}\|_2^2.
\label{eq:spatial-policy}
\end{equation}
By contrast, the dispersed budget policy samples uniformly from all subsets of cardinality $\min(b,n_i)$. Random retention first draws $\rho\sim\mathrm{Unif}[0.5,1]$ and keeps $\lceil\rho n_i\rceil$ indices. The stripe stressor samples an axis and anchor, then keeps the $b$ smallest absolute coordinate distances along that axis. All rules return at least one instance.

A classifier $f_\theta$ maps an observed bag to logits. Selection-policy shift occurs when the training mixture $\pi_{\mathrm{tr}}(p)$ and deployment mixture $\pi_{\mathrm{te}}(p)$ differ while the underlying case is unchanged. This is a structured form of input shift: the selector can induce missing-not-at-random instances and change bag cardinality, coverage, or both. Ordinary i.i.d. augmentation covers it only when its augmentation distribution matches the relevant policy family.

More formally, the risk of a fixed policy and that of a policy mixture are
\begin{equation}
 \begin{aligned}
 R_p(\theta)&=\mathbb E_{(S,y)}\mathbb E_{B\sim A_0(S)}
 \mathbb E_{B^p\sim q_p(\cdot\mid B)}
 [\ell(f_\theta(B^p),y)],\\
 R_\pi(\theta)&=\sum_{p\in\mathcal P}\pi(p)R_p(\theta).
 \end{aligned}
 \label{eq:policy-risk}
\end{equation}
The risk for a fixed policy averages randomness in cases, reference extraction, and selectors; the mixture risk then averages those risks under a deployment mixture. Thus $\pi_{\mathrm{tr}}\ne\pi_{\mathrm{te}}$ can change risk even when the case distribution and predictor are fixed. Equation~\ref{eq:policy-risk} defines the population object; our evaluation below estimates metric response within each case.

For a metric $M$ for which higher values are better, paired evaluation reports
\begin{equation}
    \Delta_M(p)=M(p_0)-M(p),
    \label{eq:degradation}
\end{equation}
where $p_0$ returns the reference bag. The same slides and matched selector draws are used across methods. We report the response curve rather than only a worst point because random thinning, fixed budgets, and localized coverage can produce different behavior. Degradation alone can be reduced by weakening the reference prediction. We therefore require a robustness claim to report both reference and restricted metric levels, and treat lower $\Delta_M$ without level preservation as insufficient.

Our training family contains random retention with keep ratio in $[0.5,1]$, a uniformly sampled budget of at most 128 patches, and a spatial policy that keeps the 128 patches nearest a random observed coordinate. Evaluation adds fixed 75\% and 50\% retention. A stripe stressor keeps the 128 patches nearest a random line and tests whether findings are specific to a compact window. Spatial and stripe policies serve as reproducible stressors based on local coordinates.

\section{Matched Policy Training Controls}

Every control shares one predictor across views. Denote the reference loss by $\ell_i^0=\ell(f_\theta(B_i^{p_0}),y_i)$ and a restricted loss by $\ell_i^p$. Average matched exposure minimizes
\begin{equation}
 \mathcal L_{\mathrm{avg}}=\frac1N\sum_i\ell_i^0+
 \frac{\alpha}{N|\mathcal P'|}\sum_i\sum_{p\in\mathcal P'}\ell_i^p,
 \label{eq:average}
\end{equation}
where $\mathcal P'=\mathcal P\setminus\{p_0\}$ and $\alpha=0.5$.

The worst-policy risk control replaces the average restricted risk with
\begin{equation}
 \mathcal L_{\mathrm{worst}}=\frac1N\sum_i\ell_i^0+
 \frac{\alpha}{N}\sum_i\max_{p\in\mathcal P'}\ell_i^p.
 \label{eq:worst}
\end{equation}
Equations~\ref{eq:average} and~\ref{eq:worst} use the same weight on restricted views $\alpha$ and differ only in the reduction within each case: uniform mean versus maximum. The sensitivity analysis at fixed weights tests this separation. The maximum is over policies for the same case, not over unrelated cohort groups. The reference term discourages robustness from being obtained solely by lowering reference performance; it is a penalty, not a hard constraint. The selector is discrete: gradients pass through the selected embeddings and predictor but not through selection.

Unless stated otherwise, $f_\theta$ is a shared gated attention MIL classifier~\cite{ilse2018attention}; policy views change only the index set $I_p$. The standard attention equations and architecture dimensions are given in the supplement so that the main notation remains specific to selection policy.

The control matrix separates four possible sources of improvement. Training on reference views only and training with random deletion test generic regularization; spatial exposure alone and matched exposure test policy experience; a sweep over $\alpha\in\{0.25,0.5,1,1.5,2\}$ tests weight on restricted views; KL consistency, the policy maximum, and an objective that adds the average and maximum test the reduction. One global $\alpha$ is selected by mean QWK over five validation policies. Because the test sensitivity curve had been inspected before this validation rule was fixed, we treat the sweep as diagnostic rather than confirmatory and report all five values in the supplement.

Algorithm~\ref{alg:policy-training} gives the shared implementation. A stable hash generates policy draws, and every control receives the same evaluation bags. The reduction operates across policies for one case before losses are averaged across cases. This ordering distinguishes the policy maximum from a robust objective over batches or providers.

\begin{algorithm}[tb]
\caption{Matched policy training with reduction within each case}
\label{alg:policy-training}
\textbf{Input}: minibatch $\{(B_i,y_i,\mathrm{id}_i)\}_{i=1}^{m}$; restricted
family $\mathcal P'$; weight $\alpha$; reduction $r\in\{\mathrm{mean},\mathrm{max}\}$\\
\textbf{Output}: updated shared predictor parameters $\theta$
\begin{algorithmic}[1]
\FOR{$i=1,\ldots,m$}
  \STATE $L_i^0\leftarrow\ell(f_\theta(B_i),y_i)$
  \FORALL{$p\in\mathcal P'$}
    \STATE $\omega_{ip}\leftarrow H(\mathrm{seed},\mathrm{epoch},\mathrm{id}_i,p)$
    \STATE $B_i^p\leftarrow\{B_i[j]:j\in I_p(B_i;\omega_{ip})\}$
    \STATE $L_i^p\leftarrow\ell(f_\theta(B_i^p),y_i)$
  \ENDFOR
  \STATE $L_i^{\mathrm{res}}\leftarrow r(\{L_i^p:p\in\mathcal P'\})$
\ENDFOR
\STATE $\mathcal L\leftarrow m^{-1}\sum_i(L_i^0+\alpha L_i^{\mathrm{res}})$
\STATE Update $\theta$ with $\nabla_\theta\mathcal L$
\end{algorithmic}
\end{algorithm}

With $|\mathcal P'|=3$, each training step evaluates one reference and three restricted bags from cached embeddings. At inference, a deduplicated union of $K$ windows uses one MIL pass, while regional prediction averaging uses $K$ passes. We do not evaluate hardware-specific timing.

\section{Experiments}

\paragraph{Data and metrics.}
PANDA provides 9,555 prostate WSIs with ISUP grades 0--5 \cite{bulten2022panda}. We use the released UNI2-h feature archives with 1,536 dimensions: patches are $256\times256$ pixels at $20\times$ magnification; the release does not specify its upstream rule for tissue selection. Slides are split into 7,647/954/954 train/validation/test slides; test providers are held fixed. A deterministic sampler caps the reference pool at 512 patches. An uncapped check across three seeds changes reference view QWK by at most 0.09 points (supplement). We report quadratic weighted kappa (QWK)~\cite{cohen1968weighted}, whose native range is $[-1,1]$, after multiplication by 100 for readability, together with balanced accuracy. Thus, for example, 94.29 denotes a native QWK of 0.9429.

The lung task distinguishes LUAD from LUSC. TCGA~\cite{weinstein2013tcga} contains 1,042 primary tumor slides from 946 patients; splits grouped by patient contain 842/101/99 train/validation/internal test slides. Untouched CPTAC-LUAD and CPTAC-LSCC~\cite{cptac2018luad,cptac2018lscc} provide 606 and 523 slides from 225 and 210 patients after metadata QC requiring tumor tissue and an acceptable tumor segment. A cache fixed across seeds draws at most 512 reference patches from raw bags with median size 7,945. Slide probabilities are averaged within patients before AUROC and balanced accuracy. CPTAC is never used for tuning or model selection.

CAMELYON16~\cite{bejnordi2017diagnostic} provides 270 development and 129 official test lymph node WSIs for metastasis detection at slide level. We split the development set once with seed 3407 into 216/54 train/validation slides and hold it fixed across model seeds to isolate model and selector variability; the untouched test set contains 80 normal, 22 macro-, and 27 micrometastatic slides. Frozen UNI2-h features with 1,536 dimensions are extracted from $256\times256$ pixel patches without overlap at $0.5\,\mu$m/pixel (approximately $20\times$), resized to $224\times224$ for the encoder. A thumbnail HSV/Otsu tissue mask followed by morphology and component filtering retains patches with at least 50\% tissue. Reference bags contain all retained patches (median 5,137; maximum 23,759), without a cap. Official lesion XML is never used for training, checkpoint selection, or policy construction; it is reserved for retrospective analysis of evidence retention. One positive test slide with a macrometastasis but without XML remains in classification metrics and is excluded only from that mechanism analysis.

\paragraph{Human Evidence Scope Test.}
Two hospital datasets pair each diagnosis with a human binary judgment of whether an image contains visible evidence for that diagnosis. Among 1,981/3,981 quality-controlled H1/H2 cases, 202/242 held-out cases contain both images labeled as containing evidence and images labeled as lacking evidence, which supports a comparison of equal cardinality within each case. Separate classifiers for each hospital use archived UNI v1 features and evaluation grouped by case; the primary estimand is the paired change in the probability assigned to the true class. The judgments define semantic selectors at test time, not training targets or consensus ground truth for diagnostic sufficiency. Aggregate results, evaluation code, split generation logic, and the annotation protocol will be released; image access remains subject to institutional governance. The complete evaluation protocol is given in the supplement.

\paragraph{Training and statistics.}
The main objective controls use seeds 3407, 2027, 17, 101, and 911. PANDA models train for eight epochs and NSCLC and CAMELYON16 models for twenty with AdamW, learning rate $10^{-4}$, weight decay $10^{-4}$, mixed precision, and checkpoint selection using validation data only. All controls use identical frozen features, splits, class weights, view definitions, and evaluation bags. We report mean $\pm$ sample standard deviation across model seeds and use ten fresh selector draws per seed where noted. Paired bootstrap intervals respect patient or provider and grade strata. Seed/draw cells share models and test slides and are therefore nested repeated measurements, not independent experimental replications. Complete estimands are given in the supplement. CAMELYON16 metrics at the operating point use the fixed probability threshold 0.5.

\paragraph{Architectures and fairness.}
The full control matrix uses gated Attention MIL~\cite{ilse2018attention}. Matched average and policy-wise worst comparisons for CLAM-SB, DSMIL, and TransMIL serve as architecture scope probes; implementation details and full metric levels are reported in the supplement.

\section{Results}

\subsection{Equal Patch Counts Preserve Unequal Evidence}

\begin{figure*}[t]
\centering
\includegraphics[width=\textwidth]{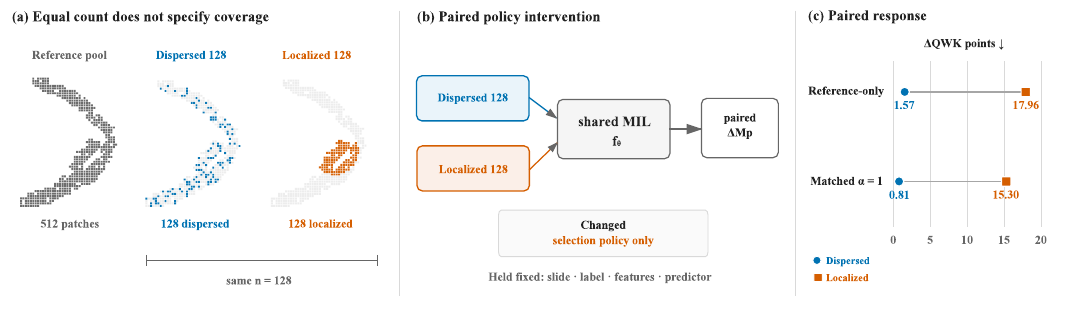}
\caption{BagShift measures selector response by paired intervention. (a) Exact PANDA coordinates for the reference pool and equally sized dispersed and localized bags; the footprints depict sampling geometry rather than histology. (b) The intervention changes only the selector while holding the slide, label, features, and predictor fixed. (c) Five-seed mean $\Delta$QWK for reference-only and validation-selected matched exposure ($\alpha=1$); exposure reduces both responses but leaves a large geometry gap.}
\label{fig:panda-protocol}
\end{figure*}

\begin{table*}[t]
\centering
\caption{PANDA test QWK on the $\times 100$ scale (mean $\pm$ SD across five seeds). The controls separate deletion, exposure, weight on restricted views, consistency, and reduction within each case. Higher absolute QWK and lower degradation are better; $\dagger$ marks the operating point selected on validation data.}
\label{tab:objective}
\footnotesize
\setlength{\tabcolsep}{3.2pt}
\begin{tabular}{llccccc}
\toprule
& & \multicolumn{3}{c}{Absolute QWK $\uparrow$}
& \multicolumn{2}{c}{Degradation (Ref.$-$policy) $\downarrow$} \\
\cmidrule(lr){3-5}\cmidrule(lr){6-7}
Control family & Training control & Ref. & Budget-128 & Spatial-128
& $\Delta$ Budget & $\Delta$ Spatial \\
\midrule
Baseline & Reference only & 94.13$\pm$0.34 & 92.56$\pm$0.81 & 76.17$\pm$2.29 & 1.57$\pm$0.78 & 17.96$\pm$2.38 \\
\addlinespace[2pt]
Deletion & Random deletion only & 94.26$\pm$0.27 & 93.17$\pm$0.92 & 77.23$\pm$1.47 & 1.09$\pm$1.03 & 17.03$\pm$1.57 \\
\addlinespace[2pt]
Exposure & Spatial-policy only & 93.96$\pm$0.36 & 92.88$\pm$0.96 & 79.30$\pm$1.04 & 1.08$\pm$0.73 & 14.66$\pm$1.21 \\
\addlinespace[2pt]
Exposure/weight & Matched average, $\alpha=0.5$ & 94.27$\pm$0.24 & 93.33$\pm$0.68 & 77.82$\pm$1.52 & 0.93$\pm$0.75 & 16.45$\pm$1.75 \\
Exposure/weight & \textbf{Matched average, $\alpha=1$}$^\dagger$ & \textbf{94.29$\pm$0.16} & \textbf{93.48$\pm$0.91} & \textbf{78.98$\pm$1.44} & \textbf{0.81$\pm$0.79} & \textbf{15.30$\pm$1.54} \\
Exposure/weight & Matched average, $\alpha=2$ & 94.21$\pm$0.31 & 93.31$\pm$0.96 & 79.51$\pm$1.20 & 0.90$\pm$0.93 & 14.70$\pm$1.42 \\
\addlinespace[2pt]
Consistency & Matched + reference KL & 94.35$\pm$0.24 & 93.47$\pm$0.79 & 78.54$\pm$1.29 & 0.88$\pm$0.81 & 15.82$\pm$1.49 \\
\addlinespace[2pt]
Robust reduction & Policy-wise worst risk & 94.39$\pm$0.37 & 93.30$\pm$0.96 & 79.65$\pm$1.15 & 1.09$\pm$0.64 & 14.74$\pm$1.37 \\
Robust reduction & Matched + worst risk & 94.41$\pm$0.33 & 93.46$\pm$0.91 & 79.72$\pm$1.41 & 0.95$\pm$0.58 & 14.68$\pm$1.56 \\
\bottomrule
\end{tabular}
\end{table*}

Figure~\ref{fig:panda-protocol} and Table~\ref{tab:objective} isolate the effect of geometry. The coordinate footprints visualize coverage rather than histology, while the reported response is computed over the full cohort. A predictor trained only on reference views loses 1.57 QWK points when 128 patches are drawn across the slide, but 17.96 points when the same number comes from one neighborhood. Matched training reduces both losses to 0.81 and 15.30 points at the operating point selected on validation data. Training helps, but it does not erase the difference between dispersed and localized evidence.

The control matrix explains where the mitigation comes from. Random deletion has little effect. Exposing the model to restricted views helps, and increasing their loss weight helps further. At $\alpha=2$, fixed reweighting reaches 79.51 spatial QWK, within $0.14\pm0.40$ points of policy-wise worst risk. This difference is small relative to seed variation and does not establish a distinct gain from worst-policy routing; the complete curve and paired intervals are given in the supplement. The useful intervention on PANDA is therefore policy exposure with sufficient weight, not a particular robust reduction.

The architecture probes separate the response from the training reduction. Under matched-average training at $\alpha=0.5$, spatial degradation remains $15.92\pm1.63$, $17.08\pm1.96$, and $15.58\pm1.44$ QWK points for CLAM-SB, DSMIL, and TransMIL, respectively. Replacing the average with the policy maximum changes spatial QWK by $+1.21\pm1.36$, $+0.61\pm1.27$, and $+0.08\pm1.33$ points; DSMIL simultaneously loses $0.42\pm0.18$ reference QWK. Thus the geometry response persists across these aggregators, while the effect of the reduction is architecture dependent rather than a general training recommendation.

The geometry gap persists across ten selector draws and a stripe stressor. It also has a direction: localized views shift positive ISUP grades downward, while random budget views leave expected grade nearly unchanged. The supplement reports the draw variability, architecture probes, and absolute reference levels.

\subsection{Localized Selection Misses Focal Evidence}

\begin{table*}[t]
\centering
\caption{CAMELYON16 response on the official test set (\%, mean $\pm$ SD across five seeds after averaging metrics for each draw within each seed). Budget and spatial views both contain 128 patches. Localized selection loses substantially more performance, and matched exposure does not recover it.}
\label{tab:camelyon-policy}
\small
\setlength{\tabcolsep}{4.0pt}
\begin{tabular}{lcccccccc}
\toprule
& \multicolumn{4}{c}{AUROC} & \multicolumn{4}{c}{Balanced accuracy} \\
\cmidrule(lr){2-5}\cmidrule(lr){6-9}
Training & Ref. & Budget & Spatial & Sp.$-$Bu. & Ref. & Budget & Spatial & Sp.$-$Bu. \\
\midrule
Reference only & 99.19$\pm$0.85 & 88.22$\pm$1.19 & 68.50$\pm$1.92 & $-$19.72$\pm$0.89 & 90.00$\pm$5.12 & 80.21$\pm$0.62 & 61.63$\pm$0.11 & $-$18.58$\pm$0.60 \\
Matched average & 99.16$\pm$0.55 & 88.08$\pm$1.10 & 67.38$\pm$2.02 & $-$20.70$\pm$1.61 & 92.24$\pm$5.52 & 80.47$\pm$1.01 & 61.71$\pm$0.20 & $-$18.76$\pm$0.85 \\
\bottomrule
\end{tabular}
\end{table*}

CAMELYON16 reproduces the gap at equal counts on an uncapped binary task. Table~\ref{tab:camelyon-policy} shows that spatial selection trails random budget sampling by 19.72 AUROC points after training on reference views only. The cohort-level ordering is consistent across the evaluated model seeds and selector draws; these 50 cells are nested repeated measurements rather than independent replications.

Lesion masks withheld from training reveal what the localized bags miss. A random budget retains annotated tumor in 52.6\% of micrometastatic observations; a spatial window retains it in only 10.0\%. On the reference view, micrometastasis-versus-normal AUROC is $98.5\pm1.4$, but sensitivity at the fixed 0.5 threshold is $63.7\pm16.8$\%, distinguishing strong ranking from the operating point. Sensitivity then falls to $29.9\pm6.3$\% under random budget sampling and $7.3\pm5.8$\% under spatial sampling. Across restricted views, tumor retention covaries with the probability response (Spearman $\rho=0.588\pm0.082$). Figure~\ref{fig:overview} makes this mechanism inspectable on a representative micrometastatic case. The case follows a deterministic median-gap rule specified in the supplement, whereas the retention estimates use all 27 micrometastatic slides; mask conversion and draw-level statistics are also provided there.

An operational mask cleanup probe shows the same failure to retain evidence. Keeping only the largest connected tissue component trails a random control with exactly matched cardinality by $5.62\pm1.12$ AUROC points, with a negative paired difference in all five seeds, and retains micrometastatic tumor in 74.1\% rather than 99.6\% of observations. The deficit is no longer persistent when the two largest components are retained. This exploratory policy family was specified after feasibility inspection, is not used for model selection, and is reported in full in the supplement.

This task also marks the limit of matched exposure. Spatial AUROC changes by $-1.12\pm2.82$ points, with no consistent recovery across seeds. The withheld masks reveal frequent mismatch between the label and visible evidence: a localized positive view can omit annotated tumor while inheriting the positive slide label. Matched exposure therefore asks the model to fit some positive bags without annotated tumor evidence, a supervision conflict consistent with the observed lack of recovery. This diagnostic does not establish that other selector-aware or instance-level strategies would fail.

\subsection{Semantic Evidence Selection in Two Hospitals}

Coordinate geometry is one way for a selector to change evidence. We test the broader principle on 444 held-out cases with multiple images from two hospitals, where human judgments identify whether each image visibly supports the case diagnosis. Selecting one image labeled as containing evidence raises the probability assigned to the true class by 19.65 points over selecting one image labeled as lacking evidence from the same case; both hospitals and all five model seeds show the same ordering. This provides a scope check that the same within-case principle can extend beyond coordinate-local patch selection; it is not WSI ROI validation. The annotation protocol, confidence intervals, and results for each hospital appear in the supplement.

\subsection{A Low-Response Boundary on TCGA-to-CPTAC}

\begin{table}[h]
\centering
\caption{CPTAC AUROC after patient aggregation (\%, mean $\pm$ SD across five seeds) after TCGA training. $\Delta$ is reference minus spatial AUROC.}
\label{tab:cptac-objective}
\footnotesize
\setlength{\tabcolsep}{3.3pt}
\begin{tabular}{lccc}
\toprule
Training control & Ref. & Spatial & $\Delta$ \\
\midrule
Average, $\alpha=0.5$ & 96.74$\pm$0.61 & 96.18$\pm$0.64 & 0.56$\pm$0.24 \\
Average, $\alpha=1$ & 96.71$\pm$0.44 & 96.08$\pm$0.51 & 0.63$\pm$0.39 \\
Worst policy & 97.04$\pm$0.25 & 96.50$\pm$0.29 & 0.54$\pm$0.19 \\
\bottomrule
\end{tabular}
\end{table}

The same stressor has little effect on lung subtyping. After training on TCGA, the matched model loses only $0.56\pm0.24$ AUROC points under localized selection on the untouched CPTAC cohort (Table~\ref{tab:cptac-objective}). Both classes move slightly toward the decision boundary, but the response is far smaller than on PANDA or CAMELYON16. This boundary case shows that the response is not uniformly large; differences in relative coverage, cohort construction, and task evidence prevent a causal cross-task attribution. Worst-policy routing does not change the degradation, and changes in balanced accuracy are mixed; full secondary metrics are given in the supplement.

\subsection{Aggregating Multiple Localized Views}

\begin{figure*}[t]
\centering
\includegraphics[width=\textwidth]{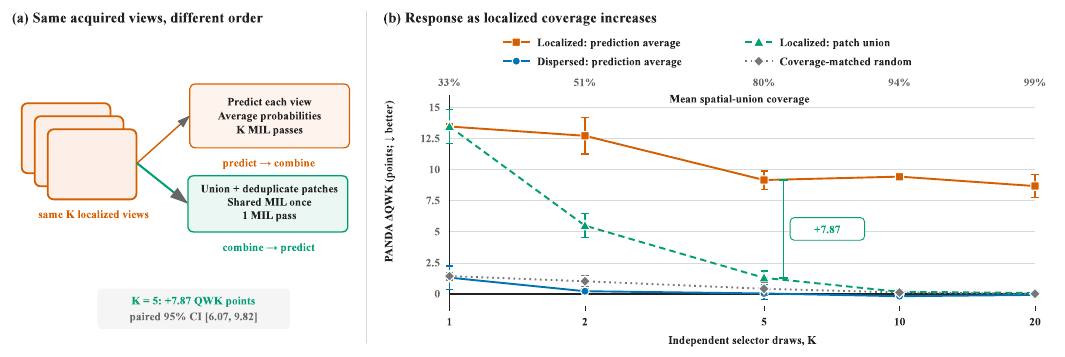}
\caption{Aggregation order changes the response to the same acquired views. (a) Prediction averaging uses $K$ independent MIL passes; union-before-MIL deduplicates patches and uses one shared pass. (b) PANDA $\Delta$QWK versus $K$ (mean $\pm$ SD across three seeds); the upper axis gives union coverage. At $K=5$, union-before-MIL gains 7.87 QWK points (paired 95\% CI 6.07--9.82).}
\label{fig:aggregation-order}
\end{figure*}

Repeated localized observations raise a distinct inference question: should a system average regional predictions or aggregate their patches as one bag? Figure~\ref{fig:aggregation-order} holds the selector draws fixed and changes only this aggregation order. Dispersed prediction averaging and a random union matched in coverage control for view count and retained cardinality, separating aggregation order from the remaining selection geometry.

Let $B_i^{(1)},\ldots,B_i^{(K)}$ be the exact localized draws and $\mathbf p_\theta(B)=\operatorname{softmax}(f_\theta(B))$. The compared probability vectors are
\begin{equation}
 \widehat{\mathbf p}_i^{\mathrm{avg}}
 =\frac1K\sum_{k=1}^K\mathbf p_\theta(B_i^{(k)}),\qquad
 \widehat{\mathbf p}_i^{\mathrm{union}}
 =\mathbf p_\theta\!\left(\bigcup_{k=1}^K B_i^{(k)}\right),
 \label{eq:union}
\end{equation}
where the union removes duplicate original patch indices. Both use identical draws; only the order of set union and nonlinear MIL aggregation differs.

At $K=5$, union-before-MIL improves QWK by 7.87 points over prediction averaging (95\% CI 6.07--9.82). The gap is not a consequence of using too few windows. At $K=20$, their exact union covers 99.3\% of the PANDA reference bag and returns within 0.08 QWK points of the reference prediction; averaging the same 20 regional predictions remains 8.68 points below it. Once separate MIL passes have compressed each region, averaging cannot reconstruct the joint bag representation.

The conclusion also holds beyond the final class decision. On PANDA, union improves ranked probability score by $0.0185\pm0.0013$ and macro one-versus-rest AUROC by $1.84\pm0.11$ points. Thus the gain is not an artifact of the decision rule, although its 7.87-point QWK effect is larger than its ranking effect. On CAMELYON16, the gain in balanced accuracy is large ($18.78\pm1.55$ points), whereas the smaller AUROC gain ($1.43\pm0.72$ across seeds) has a paired bootstrap interval spanning zero ($-1.25$ to $3.83$ points). Log loss nevertheless decreases from $0.581\pm0.005$ to $0.482\pm0.020$, indicating improved probability quality even though much of the decision benefit arises at the operating point. A random union with matched cardinality still outperforms the spatial union, so correct aggregation recovers information across observed regions but does not replace broad coverage. Complete curves and intervals appear in the supplement.

\section{Discussion and Scope}

The three public tasks suggest an evidence locality hypothesis. ISUP grading depends on primary and secondary Gleason patterns~\cite{vanleenders2020isup}, and CAMELYON16 positives may contain a small metastatic focus. Localized views can remove evidence that determines the label in both settings. LUAD/LUSC morphology was less sensitive to the same stressor in the evaluated CPTAC cohort. Lesion retention supports this interpretation on CAMELYON16, but the effect sizes across tasks do not identify evidence locality in isolation. In particular, 128 patches are one quarter of the cap of 512 patches used for PANDA and CPTAC, but only about 2.5\% of the median uncapped CAMELYON16 bag. Cohort composition and aggregation by patient also differ. The comparison of budget and spatial views with the same cardinality within CAMELYON16 still isolates selection geometry at the tested budget; comparisons of severity across tasks remain descriptive.

\paragraph{Interpreting a policy response.}
A BagShift response is indexed by its reference extraction, selector family, budget, and aggregation rule; it is not an intrinsic robustness score. Equal-cardinality comparisons ask what changes when the case and compute budget are fixed but the spatial distribution of observed instances changes. They do not rank selectors universally: semantic selection may concentrate relevant tissue, whereas coordinate-local selection may omit it. CAMELYON16 adds retained lesion evidence to the response; without comparable annotations, the response remains an empirical system test rather than a causal measure of evidence sufficiency.

\paragraph{Deployment audit.}
A BagShift audit should specify the candidate pool---magnification, tissue mask, quality control, and cap---and describe selectors by cardinality and coverage. It should apply paired policies to the same cases, report absolute and degraded performance, use lesion or ROI retention when available, and state whether repeated regions are averaged as predictions or united before nonlinear aggregation. This separates compute, observation, and aggregation choices that are otherwise hidden in preprocessing.

The coordinate policies and tissue-component cleanup are reproducible stressors, not prospective validation of a clinical ROI workflow. PANDA responses are conditional on a released candidate pool whose upstream tissue-selection rule is unspecified. The CAMELYON16 mechanism analysis is retrospective and contains 27 micrometastatic slides; public experiments use frozen UNI2-h features, and H1/H2 image access is institutionally governed. A specific deployment should therefore audit its own selector with human oversight and subgroup error analysis.

\FloatBarrier
\section{Conclusion}

Whole-slide MIL is coupled not only to patch count but also to the evidence admitted by its selector. BagShift measures this coupling within cases. Equal-sized bags produce responses ranging from small on lung subtyping to large when localized views omit focal evidence. Exposure and reweighting reduce the PANDA response, whereas matched exposure does not consistently recover on CAMELYON16 when positive views often omit annotated tumor. Repeated observations introduce another system choice: union patches before nonlinear MIL rather than average separately compressed predictions. Patch count specifies computation, not observation; deployments should report evidence retention and repeated-view aggregation.


\bibliography{references}

\input{arxiv_supplement}
\end{document}

%% file: arxiv_supplement.tex
\FloatBarrier
\clearpage 
\section*{Supplementary Material}
\setcounter{secnumdepth}{1}
\setcounter{section}{0}
\setcounter{subsection}{0}
\setcounter{table}{0}
\setcounter{figure}{0}
\setcounter{equation}{0}
\renewcommand{\thesection}{S\arabic{section}}
\renewcommand{\thesubsection}{\thesection.\arabic{subsection}}
\renewcommand{\thetable}{S\arabic{table}}
\renewcommand{\thefigure}{S\arabic{figure}}
\renewcommand{\theequation}{S\arabic{equation}}

\section{Scope and Evidence Boundary}

This supplement provides detailed public-cohort diagnostics and two distinct
multi-image analyses. The first is a supervised two-hospital scope test with
human image-level visible-evidence annotations. The second is a
separate, restricted three-hospital image--report retrieval audit. Neither is
presented as WSI ROI validation.

\section{Shared Attention MIL Implementation}

The main Attention MIL classifier projects each frozen embedding as
$\mathbf h_{ij}=g(\mathbf x_{ij})$ and applies the standard gated attention
operator
\begin{equation}
\begin{aligned}
 e_{ij}&=\mathbf w^\top\!\left[
 \tanh(V\mathbf h_{ij})\odot\sigma(U\mathbf h_{ij})\right],\\
 a_{ij}&=\frac{\exp(e_{ij})}{\sum_{k\in I_p}\exp(e_{ik})},
 &\mathbf z_i^p&=\sum_{j\in I_p}a_{ij}\mathbf h_{ij}.
\end{aligned}
\end{equation}
The projected and attention widths are 256; the normalized bag embedding has
128 dimensions and feeds a linear classifier. Every policy view shares
$g,U,V,\mathbf w$ and the classifier; only $I_p$ changes.

\section{Detailed Public-Cohort Policy Diagnostics}

\subsection{PANDA Reference-Pool Sensitivity}

An earlier three-seed check evaluates the full-view level with either the
deterministic 512-patch reference pool or all stored patches (up to 2,770).
For reference-only Attention MIL, QWK is $93.56\pm0.79$ with the cap and
$93.65\pm0.75$ with all stored patches. For the corresponding policy-trained
Attention configuration, it is $94.19\pm0.23$ and $94.21\pm0.31$,
respectively. This check shows little full-view sensitivity to the cap in those
checkpoints; because restricted policies were not recomputed from every
all-stored bag, it does not establish cap invariance of the complete response
curve.

\begin{figure*}[t]
\centering
\includegraphics[width=\textwidth]{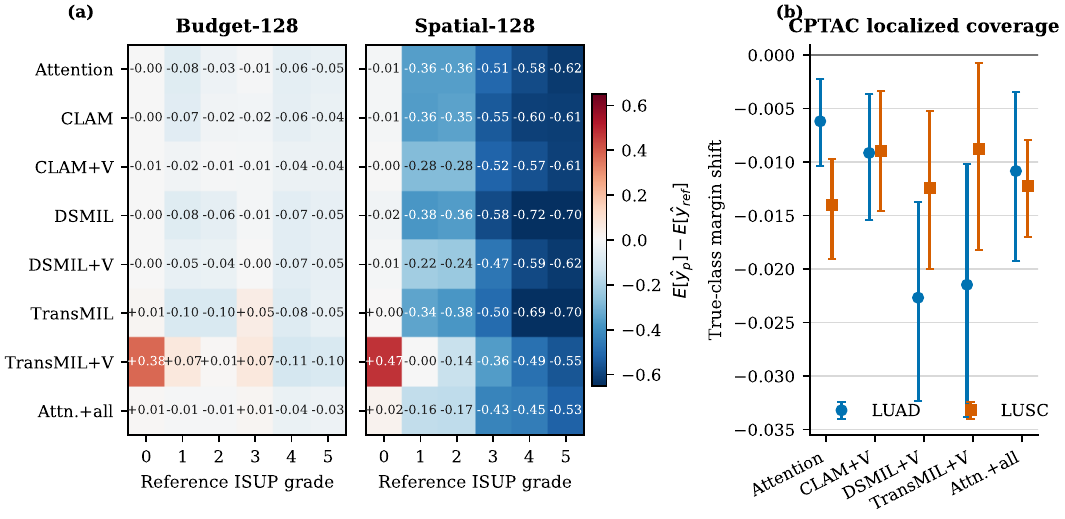}
\caption{Detailed selector diagnostics. (a) Three-seed mean expected-grade
shifts for budget and spatial selectors on PANDA. (b) CPTAC spatial-window
true-class margin shifts with conditional patient-bootstrap intervals. Negative
values denote movement away from the true class.}
\label{fig:detailed-geometry}
\end{figure*}

Figure~\ref{fig:detailed-geometry} retains the complete eight-configuration
PANDA heatmaps and the external CPTAC margin response omitted from the main
paper. Repeated-selector aggregation is shown only once, in main Figure~3.

\subsection{Auditable Selector Cases}

\begin{figure*}[t]
\centering
\includegraphics[width=\textwidth]{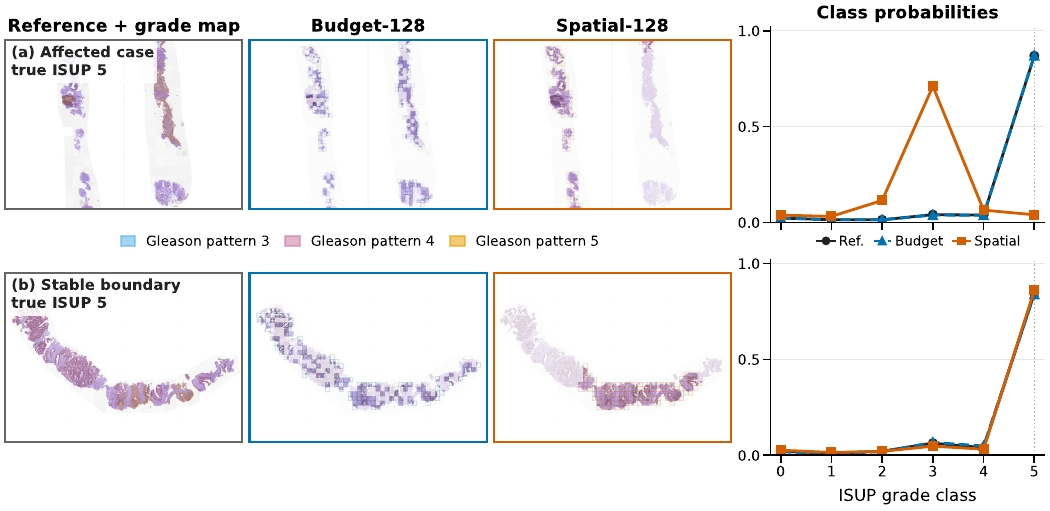}
\caption{Two deterministically selected PANDA cases on level-1 H\&E. The reference
column overlays the retrospective Radboud grade mask; the middle columns show the
exact 128-patch views. The tall first slide is folded into upper and lower halves
without rescaling. All panels preserve tissue aspect ratio; white padding only
standardizes the display cells. Masks were unavailable to model and selector
development; predictions use the unchanged validation-selected Attention control.
In the affected case, reference and Budget-128 distributions nearly coincide
while Spatial-128 shifts to grade 3; in the stable case, all three distributions
nearly coincide by the frozen case-selection rule.}
\label{fig:s-selector-cases}
\end{figure*}

Figure~\ref{fig:s-selector-cases} makes two individual selection outcomes
inspectable without treating them as population estimates. The cases were
chosen by a frozen rule before rendering. Among test slides with label 3--5 and
at least 256 stored patches, the affected case is closest to the median
expected-grade shift among slides whose spatial hard prediction falls below the
reference prediction. The boundary case is the grade-5 slide with unchanged
hard prediction and the smallest absolute expected-grade shift. Predictions and
selectors use the validation-selected Attention control, model seed 3407, and
selector draw 0. Public aspect-ratio-preserving level-1 H\&E images and
level-2 grade masks are added only
for retrospective visualization; neither source entered training, checkpoint
selection, selector construction, or the frozen case-selection rule.

\subsection{Architecture-Matched External Transfer}

Table~\ref{tab:s-external} gives the cross-architecture control summarized in
the main paper. These are three-seed patient-level means. Matched exposure is
neutral-to-positive for CLAM-SB, approximately neutral in DSMIL AUROC, and
negative for TransMIL. The TransMIL loss is already present on the reference
view and is therefore not an extra spatial-policy degradation.

\begin{table}[t]
\centering
\small
\setlength{\tabcolsep}{3.0pt}
\begin{tabular}{lccc}
\toprule
Model & Reference & Spatial 128 & $\Delta$ \\
\midrule
CLAM-SB & 96.40$\pm$1.36 & 95.51$\pm$1.93 & 0.90$\pm$0.58 \\
CLAM-SB + views & \textbf{97.13}$\pm$0.07 & \textbf{96.56}$\pm$0.36 & \textbf{0.56}$\pm$0.29 \\
DSMIL & 96.48$\pm$0.23 & 95.40$\pm$0.08 & 1.08$\pm$0.16 \\
DSMIL + views & 96.26$\pm$0.44 & 95.36$\pm$0.48 & 0.91$\pm$0.28 \\
TransMIL & 96.85$\pm$0.49 & 96.11$\pm$0.50 & 0.73$\pm$0.21 \\
TransMIL + views & 95.18$\pm$0.27 & 94.18$\pm$0.45 & 1.00$\pm$0.18 \\
\bottomrule
\end{tabular}
\caption{CPTAC patient AUROC after TCGA training (\%).}
\label{tab:s-external}
\end{table}

Disabling PPEG preserves the PANDA exposure benefit: spatial degradation falls
from 18.25 to 16.79 QWK points. On CPTAC, however, the reference view-training
effect becomes $+0.42/-2.05/+1.30$ AUROC points across seeds, and the spatial
effect becomes $+0.42/-2.43/+1.28$. Removing PPEG therefore makes external
accuracy less stable rather than resolving the failure. Because this control
does not restore physical coordinates, it cannot identify pseudo-grid geometry
as the cause.

\subsection{Five-Seed Objective Diagnostics}

Tables~\ref{tab:s-objective-xarch} and~\ref{tab:s-objective-cptac} retain the
objective comparisons removed from the main paper. Their average-risk rows use
$\alpha=0.5$. The Attention sensitivity curve shows that a larger fixed weight
can close the worst-policy gap, but equivalent weight sweeps were not run for
these architectures or NSCLC. The tables therefore mix routing with restricted-
view weight and cannot estimate the isolated contribution of the maximum.

\begin{table}[t]
\centering
\small
\setlength{\tabcolsep}{3.2pt}
\begin{tabular}{llccc}
\toprule
Model & Objective & Reference & Spatial & $\Delta$ \\
\midrule
CLAM-SB & Average & 93.44$\pm$0.60 & 77.52$\pm$1.84 & 15.92$\pm$1.63 \\
CLAM-SB & Worst & 93.80$\pm$0.53 & 78.72$\pm$1.40 & 15.08$\pm$1.40 \\
DSMIL & Average & 94.44$\pm$0.61 & 77.36$\pm$2.04 & 17.08$\pm$1.96 \\
DSMIL & Worst & 94.03$\pm$0.52 & 77.97$\pm$0.96 & 16.05$\pm$1.05 \\
TransMIL & Average & 94.73$\pm$0.27 & 79.15$\pm$1.64 & 15.58$\pm$1.44 \\
TransMIL & Worst & 94.54$\pm$0.58 & 79.23$\pm$1.57 & 15.31$\pm$1.49 \\
\bottomrule
\end{tabular}
\caption{Five-seed PANDA QWK on the $\times 100$ scale for matched-average $\alpha=0.5$ and
worst-policy objectives using identical views.}
\label{tab:s-objective-xarch}
\end{table}

At this fixed weight, CLAM-SB spatial QWK changes by $+1.21\pm1.36$, DSMIL
spatial QWK changes by $+0.61\pm1.27$ while reference QWK falls by
$0.42\pm0.18$, and TransMIL spatial QWK changes by $+0.08\pm1.33$. These are
scope diagnostics, not evidence that routing transfers across aggregators.

\begin{table*}[t]
\centering
\small
\setlength{\tabcolsep}{4.0pt}
\begin{tabular}{lccc}
\toprule
Control & Ref. AUC/BAcc & Spatial AUC/BAcc & $\Delta$ \\
\midrule
Average, $\alpha=0.5$ & 96.74$\pm$0.61/90.49$\pm$1.52 & 96.18$\pm$0.64/90.13$\pm$1.61 & 0.56$\pm$0.24 \\
Average, $\alpha=1$ & 96.71$\pm$0.44/89.67$\pm$1.34 & 96.08$\pm$0.51/89.40$\pm$1.51 & 0.63$\pm$0.39 \\
Worst policy & 97.04$\pm$0.25/90.57$\pm$1.28 & 96.50$\pm$0.29/89.66$\pm$1.52 & 0.54$\pm$0.19 \\
\bottomrule
\end{tabular}
\caption{Untouched CPTAC patient results after TCGA training (\%, five-seed
mean $\pm$ SD). Cells report AUROC/BAcc; $\Delta$ is reference minus spatial
AUROC.}
\label{tab:s-objective-cptac}
\end{table*}

Worst policy raises reference and spatial AUROC over $\alpha=0.5$ by
$0.30\pm0.48$ and $0.32\pm0.42$, but AUROC degradation is unchanged and
balanced accuracy is mixed. No NSCLC alpha sweep was run.

\subsection{Earlier CVaR and EMA Diagnostics}

The initial Attention configuration added environment--policy positive-regret
CVaR and a reference-view EMA teacher. Table~\ref{tab:s-components} records this
earlier three-seed diagnostic, which differs from the later five-seed objective
matrix in both configuration and seed set. Its absolute values therefore do not
rank methods against the main-paper controls. It motivated the simpler matrix:
PANDA's CVaR increment is reproduced by policy-wise worst loss without
environment construction, while EMA and CVaR have no uniform cross-task gain.

\begin{table}[t]
\centering
\small
\setlength{\tabcolsep}{3.0pt}
\begin{tabular}{lccc}
\toprule
Variant & PANDA & CPTAC ref. & CPTAC spatial \\
\midrule
Attention & 74.88 & 96.73 / 89.73 & 95.72 / 88.53 \\
+ policy views & 77.20 & 96.82 / 90.08 & 96.00 / 89.31 \\
+ CVaR & \textbf{79.73} & 96.83 / 90.13 & 96.13 / 89.37 \\
+ EMA loss & 79.68 & \textbf{97.10 / 90.96} & \textbf{96.43 / 90.25} \\
\bottomrule
\end{tabular}
\caption{Earlier Attention component means on $\times 100$ scales. PANDA reports
spatial QWK; CPTAC reports patient AUROC / balanced accuracy.}
\label{tab:s-components}
\end{table}

These ablations were designed after inspecting the initial PANDA result. Their
higher spatial point estimates are therefore exploratory, not untouched
estimates of superiority. The new five-seed controls use neither environment
labels nor an EMA teacher and provide the comparison reported in the main
paper.

\subsection{Fixed Restricted-Weight Sweep}

\begin{figure}[t]
\centering
\includegraphics[width=\columnwidth]{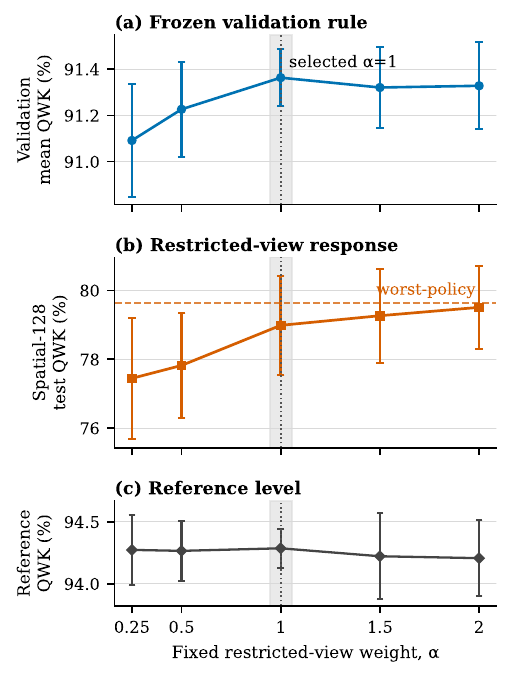}
\caption{Fixed-weight sensitivity. (a) One global $\alpha$ is selected by mean
QWK over five validation policies, five model seeds, and ten selector draws.
(b) Spatial-128 test response; the dashed line marks worst-policy performance.
(c) Absolute reference QWK across the same weights. Error bars are SD across
seeds; the shaded band and dotted line mark the selected $\alpha$.}
\label{fig:s-alpha-response}
\end{figure}

Table~\ref{tab:s-alpha} reports the complete five-seed curve for fixed
restricted-view weight $\alpha$. A retrospective rule selects one global
$\alpha$ by mean QWK over five validation policies, five seeds, and ten draws.
It selects $\alpha=1$; $\alpha=2$ has the highest validation spatial QWK but a
slightly lower policy mean. Because the test curve was inspected first, this is
a fairness check rather than untouched hyperparameter confirmation; overlapping
seed variation does not identify a unique optimum.

\begin{table*}[t]
\centering
\small
\setlength{\tabcolsep}{5.0pt}
\begin{tabular}{lcccc}
\toprule
Control & Val. mean & Reference & Spatial & $\Delta$ \\
\midrule
Average, $\alpha=0.25$ & 91.09$\pm$0.24 & 94.27$\pm$0.29 & 77.44$\pm$1.76 & 16.83$\pm$1.99 \\
Average, $\alpha=0.50$ & 91.23$\pm$0.20 & 94.27$\pm$0.24 & 77.82$\pm$1.52 & 16.45$\pm$1.75 \\
Average, $\alpha=1.00$ & \textbf{91.36}$\pm$0.12 & 94.29$\pm$0.16 & 78.98$\pm$1.44 & 15.30$\pm$1.54 \\
Average, $\alpha=1.50$ & 91.32$\pm$0.18 & 94.22$\pm$0.35 & 79.26$\pm$1.37 & 14.96$\pm$1.69 \\
Average, $\alpha=2.00$ & 91.33$\pm$0.19 & 94.21$\pm$0.31 & 79.51$\pm$1.20 & 14.70$\pm$1.42 \\
Worst policy & -- & 94.39$\pm$0.37 & 79.65$\pm$1.15 & 14.74$\pm$1.37 \\
\bottomrule
\end{tabular}
\caption{Complete fixed-weight QWK curve on the $\times 100$ scale (five-seed
mean $\pm$ SD). Validation is the mean QWK over five policies; remaining
columns are PANDA test results.}
\label{tab:s-alpha}
\end{table*}

The paired worst-minus-$\alpha=2$ spatial difference is $+0.14\pm0.40$ QWK
points; four of five seeds are positive, but all five provider-by-grade
bootstrap intervals include zero. The evidence does not show a clear
worst-policy advantage beyond sufficiently strong fixed reweighting.

\begin{table*}[t]
\centering
\small
\setlength{\tabcolsep}{5.0pt}
\begin{tabular}{lccccc}
\toprule
Control & Ref. & Sp. $\Delta$ & Draw SD & Max $\Delta$ & Stripe $\Delta$ \\
\midrule
Spatial-policy only & 93.96 & \textbf{14.01} & 1.41 & 16.13 & \textbf{11.88} \\
Validation-selected $\alpha=1$ & 94.29 & 14.88 & 1.36 & 17.01 & 12.58 \\
Sensitivity endpoint $\alpha=2$ & 94.21 & 14.22 & 1.44 & 16.36 & 12.05 \\
Policy-wise worst & \textbf{94.39} & 14.03 & \textbf{1.33} & \textbf{16.12} & 11.95 \\
\bottomrule
\end{tabular}
\caption{PANDA selector variability over five model seeds and ten draws per
seed. Ref. is reference QWK; remaining policy values are QWK degradation
points, and SD is across selector draws.}
\label{tab:s-draws}
\end{table*}

\subsection{Cross-Task Multi-Window Mechanism Check}

Table~\ref{tab:s-panda-k} gives the complete PANDA response curve underlying
main Figure~3. Every row reuses the same selector draws across aggregation
rules; coverage is the fraction of the reference bag represented by the exact
localized union.

\begin{table*}[t]
\centering
\small
\setlength{\tabcolsep}{3.4pt}
\begin{tabular}{rrrrrr}
\toprule
$K$ & Coverage & \shortstack{Localized prediction\\average} &
\shortstack{Localized patch\\union} & \shortstack{Dispersed prediction\\average} &
\shortstack{Coverage-matched\\random} \\
\midrule
1  & 32.8\% & $13.47\pm1.38$ & $13.47\pm1.38$ & $1.31\pm0.94$ & $1.43\pm0.31$ \\
2  & 51.4\% & $12.72\pm1.46$ & $5.50\pm0.98$  & $0.22\pm0.23$ & $1.02\pm0.44$ \\
5  & 79.9\% & $9.16\pm0.73$  & $1.29\pm0.56$  & $0.04\pm0.51$ & $0.42\pm0.51$ \\
10 & 94.4\% & $9.43\pm0.25$  & $0.18\pm0.16$  & $-0.18\pm0.15$ & $0.12\pm0.27$ \\
20 & 99.3\% & $8.68\pm0.95$  & $0.08\pm0.12$  & $-0.08\pm0.10$ & $0.01\pm0.03$ \\
\bottomrule
\end{tabular}
\caption{PANDA QWK degradation in points (three-seed mean $\pm$ SD) as the
number of localized observations increases. Lower is better; negative values
indicate a small improvement over the reference view. At $K=5$, localized
union improves over localized prediction averaging by 7.87 points (paired
95\% CI 6.07--9.82).}
\label{tab:s-panda-k}
\end{table*}

Table~\ref{tab:s-mechanism} uses the same 20 selector draws for regional
prediction averaging and exact patch union. Above-chance balanced-accuracy
skill is normalized to the task-specific chance level. Direction is expected
grade change divided by five for PANDA and true-class margin change for CPTAC.
PANDA retains substantial directional loss after averaging 20 localized
predictions, whereas exact union reconstructs the reference response. The
CPTAC effect is much smaller. This supports persistent local coverage and
aggregation order as a computational explanation; it cannot identify tissue-
level biological causation.

\begin{table}[t]
\centering
\small
\setlength{\tabcolsep}{3.0pt}
\begin{tabular}{llrr}
\toprule
Task & Strategy & Skill loss & Directional shift \\
\midrule
PANDA & Spatial average & 16.45\% & $-0.0408$ \\
PANDA & Spatial union & 0.08\% & $+0.0000$ \\
CPTAC & Spatial average & 0.99\% & $-0.0097$ \\
CPTAC & Spatial union & 0.00\% & $-0.0000$ \\
\bottomrule
\end{tabular}
\caption{Twenty-draw mechanism check. Skill loss and directional shift compare
the restricted strategy with its reference prediction.}
\label{tab:s-mechanism}
\end{table}

The conditional 95\% interval for PANDA spatial-average skill loss is
13.22--19.61\%; the CPTAC interval crosses zero. Normalized within-case
dispersion beyond the random-budget control is 0.0466 on PANDA and 0.0150 on
CPTAC, a 3.10-fold ratio.

\subsection{Held-Out Policy Family}

This experiment reuses the earlier Attention baseline in
Table~\ref{tab:s-components}, but unlike its ``+ policy views'' row, it excludes
spatial views from training. Random-keep and budget exposure raise spatial QWK
from $74.88\pm1.60$ to $76.50\pm1.62$. The paired seed changes
are $+1.94/+1.55/+1.39$, with provider-by-grade bootstrap intervals
$[+0.95,+3.08]$, $[+0.59,+2.55]$, and $[+0.42,+2.62]$. Adding the earlier CVaR
term yields $76.64\pm1.45$; one paired seed is negative and all three
incremental intervals include zero. Thus tested policy exposure transfers
partly across families, whereas the CVaR increment does not.

\section{CAMELYON16 Focal-Evidence Validation}

\subsection{Protocol and Integrity Checks}

CAMELYON16 uses the official 270-slide development set and untouched 129-slide
test set. A seed-3407 stratified split assigns 216/54 development slides to
training/validation; the test set contains 80 normal, 22 macro-, and 27
micrometastatic slides. UNI2-h features are 1,536-dimensional and come from
non-overlapping $256\times256$ patches at $0.5\,\mu$m/pixel, resized to
$224\times224$ for the encoder. Thumbnail HSV-saturation Otsu masking, gray
rejection, morphology, connected-component filtering, and a 50\% tissue
threshold define the reference bag. The feature audit covers all 399 slides:
retained patch counts range from 112 to 23,759 (median 5,137), and no effective
reference cap is applied. All available annotated positives retain at least
one tumor-overlap patch. One official macro-positive test slide has no XML;
it remains in classification metrics and is excluded only from lesion-retention
analysis.

For lesion analysis, XML coordinates are scaled from level-0 pixels to the
tissue-mask raster. Positive polygons are filled first and exclusion polygons
are then removed, making the result independent of XML member order. Each
retained patch is mapped through its level-0 footprint and counted as
tumor-positive when the mean binary mask value over that footprint is greater
than zero (any positive-polygon overlap). Lesion masks are stored only as
per-patch overlap fractions and are never read by training or selection code.

Reference-only and matched-average Attention MIL use seeds 3407, 2027, 17,
101, and 911. Each model trains for 20 epochs with validation-AUROC checkpoint
selection. Evaluation uses ten paired selector draws for full, 75\% and 50\%
random retention, random-budget-128, and spatial-window-128 views. Values in
Table~\ref{tab:s-camelyon-policy} first average draws within a seed and then
report mean $\pm$ sample SD across seeds.
Balanced accuracy and stage sensitivity use the fixed probability threshold
0.5. The 50 seed/draw cells are nested repeated measurements because they share
five trained models and the same official test slides; they are not treated as
50 independent experiments.
For reference-only spatial views, draw-level metrics averaged within each seed
give $23.59\pm0.18$\% sensitivity, $99.68\pm0.17$\% specificity, and a
$9.16\pm0.14$\% positive-prediction rate. These values explain the small
between-seed variance in spatial balanced accuracy as a stable bias toward the
negative class.

\begin{table*}[t]
\centering
\small
\setlength{\tabcolsep}{4.0pt}
\begin{tabular}{llccc}
\toprule
Training & Policy & AUROC & AUPRC & Balanced accuracy \\
\midrule
Reference only & Full & $0.9919\pm0.0085$ & $0.9898\pm0.0092$ & $0.9000\pm0.0512$ \\
& Random keep 75\% & $0.9902\pm0.0095$ & $0.9884\pm0.0099$ & $0.8980\pm0.0404$ \\
& Random keep 50\% & $0.9844\pm0.0084$ & $0.9832\pm0.0081$ & $0.8943\pm0.0352$ \\
& Random budget 128 & $0.8822\pm0.0119$ & $0.8828\pm0.0079$ & $0.8021\pm0.0062$ \\
& Spatial window 128 & $0.6850\pm0.0192$ & $0.6647\pm0.0150$ & $0.6163\pm0.0011$ \\
\midrule
Matched average & Full & $0.9916\pm0.0055$ & $0.9894\pm0.0068$ & $0.9224\pm0.0552$ \\
& Random keep 75\% & $0.9902\pm0.0067$ & $0.9881\pm0.0079$ & $0.9218\pm0.0515$ \\
& Random keep 50\% & $0.9852\pm0.0088$ & $0.9837\pm0.0091$ & $0.9170\pm0.0406$ \\
& Random budget 128 & $0.8808\pm0.0110$ & $0.8827\pm0.0099$ & $0.8047\pm0.0101$ \\
& Spatial window 128 & $0.6738\pm0.0202$ & $0.6523\pm0.0216$ & $0.6171\pm0.0020$ \\
\bottomrule
\end{tabular}
\caption{CAMELYON16 official-test policy response. Metrics are proportions;
budget and spatial views contain 128 patches.}
\label{tab:s-camelyon-policy}
\end{table*}

Spatial-minus-budget AUROC is $-0.1972\pm0.0089$ for reference-only and
$-0.2070\pm0.0161$ for matched training across seed-wise means; all 50
seed/draw differences are negative. Label-stratified paired bootstraps on mean
predictions give $-0.1357$ (95\% CI $-0.2194$ to $-0.0617$) and $-0.1454$
($-0.2278$ to $-0.0750$), respectively. Matched-minus-reference spatial AUROC
is $-0.0112\pm0.0282$ (one of five seeds positive), while balanced accuracy
changes by $+0.0008\pm0.0026$.

\subsection{Stage and Lesion Retention}

Table~\ref{tab:s-camelyon-stage} stratifies positive slides by official stage.
Macro/micro AUROC compares that stage with all 80 normal slides. Lesion XML is
used only after inference to ask whether a selected bag contains at least one
patch overlapping annotated tumor.

\begin{table*}[t]
\centering
\small
\setlength{\tabcolsep}{3.6pt}
\begin{tabular}{lllccc}
\toprule
Training & View & Stage & Sensitivity & AUROC vs. normal & Mean positive probability \\
\midrule
Reference only & Full & Macro & $1.000\pm0.000$ & $1.000\pm0.000$ & $0.729$ \\
& Full & Micro & $0.637\pm0.168$ & $0.985\pm0.014$ & $0.568$ \\
& Budget 128 & Macro & $0.982\pm0.030$ & $0.994\pm0.011$ & $0.719$ \\
& Budget 128 & Micro & $0.299\pm0.063$ & $0.791\pm0.034$ & $0.406$ \\
& Spatial 128 & Macro & $0.435\pm0.062$ & $0.747\pm0.062$ & $0.471$ \\
& Spatial 128 & Micro & $0.073\pm0.058$ & $0.634\pm0.054$ & $0.308$ \\
\midrule
Matched average & Full & Macro & $1.000\pm0.000$ & $1.000\pm0.000$ & $0.739$ \\
& Full & Micro & $0.719\pm0.181$ & $0.985\pm0.009$ & $0.613$ \\
& Budget 128 & Macro & $0.981\pm0.031$ & $0.995\pm0.011$ & $0.729$ \\
& Budget 128 & Micro & $0.316\pm0.076$ & $0.788\pm0.033$ & $0.416$ \\
& Spatial 128 & Macro & $0.438\pm0.064$ & $0.742\pm0.058$ & $0.476$ \\
& Spatial 128 & Micro & $0.076\pm0.060$ & $0.618\pm0.058$ & $0.309$ \\
\bottomrule
\end{tabular}
\caption{CAMELYON16 stage response (mean $\pm$ SD across 50 nested seed/draw
cells). The full view is deterministic within a model seed.}
\label{tab:s-camelyon-stage}
\end{table*}

For annotated positives, budget-128 retains tumor evidence in 100.0\% of
macro and 52.6\% of micro observations; spatial-128 retains it in 42.9\% and
10.0\%. Mean Spearman correlation between tumor retention and the
restricted-minus-reference probability shift is $0.588\pm0.082$ for
reference-only spatial views and $0.615\pm0.086$ for matched spatial views. The
corresponding budget values are $0.418\pm0.061$ and $0.452\pm0.068$.

\subsection{Post-Hoc Tissue-Component Cleanup}

To test a selector closer to operational tissue-mask cleanup, we form an
8-connected graph on the complete tissue-filtered extraction grid and retain
the $K$ largest components. The grid step is read from each feature file rather
than inferred from sparse retained coordinates. For every slide and $K$, ten
random controls retain exactly the same number of patches. This analysis is
post-hoc exploratory: the component family was selected after feasibility
inspection of official-test lesion metadata. We therefore report the complete
$K\in\{1,2,3,5,10\}$ curve rather than selecting an operating point.

\begin{table*}[t]
\centering
\small
\setlength{\tabcolsep}{3.8pt}
\begin{tabular}{rrrrrr}
\toprule
$K$ & Median retained & Component AUROC & Matched-random AUROC & Difference & \shortstack{Micro tumor retained\\component / random} \\
\midrule
1  & 44.9\% & $92.02\pm1.46$ & $97.64\pm0.84$ & $-5.62\pm1.12$ & 74.1\% / 99.6\% \\
2  & 79.8\% & $99.17\pm0.61$ & $98.82\pm0.82$ & $+0.35\pm0.31$ & 88.9\% / 99.6\% \\
3  & 89.3\% & $99.14\pm0.68$ & $99.04\pm0.95$ & $+0.10\pm0.34$ & 96.3\% / 100.0\% \\
5  & 96.2\% & $99.28\pm0.62$ & $99.11\pm0.93$ & $+0.17\pm0.32$ & 100.0\% / 100.0\% \\
10 & 98.4\% & $99.22\pm0.78$ & $99.18\pm0.86$ & $+0.03\pm0.08$ & 100.0\% / 100.0\% \\
\bottomrule
\end{tabular}
\caption{Exploratory CAMELYON16 tissue-component cleanup. AUROC is in
percentage points (five-seed mean $\pm$ SD); random metrics first average ten
draws within each seed. Difference is component minus exact-cardinality random.
At $K=1$, all five seed differences are negative. A label-stratified paired
bootstrap on seed/draw-averaged slide predictions gives $-7.55$ AUROC points
(95\% CI $-14.18$ to $-2.17$).}
\label{tab:s-camelyon-components}
\end{table*}

The curve both identifies and bounds the operational risk. Aggressively keeping
only the largest tissue fragment can discard focal evidence despite a large
remaining bag. Retaining two components raises the median retained fraction
from 44.9\% to 79.8\%, after which the deficit relative to the exact-cardinality
control is no longer persistent; by five components, micro-tumor retention is
complete in these draws. Thus the large localized-window response should not be
read as a claim that routine component cleanup is generally destructive. The
result instead shows why the exact cleanup setting belongs in a deployment
audit.

\paragraph{Figure 1 case-selection protocol.}
Eligible official-test slides had a micrometastatic label, annotated tumor
retained by the fixed dispersed-128 draw but not by the fixed localized-128
draw, and five-seed mean predictions on opposite sides of the 0.5 operating
threshold. Among the five eligible slides, we selected the slide whose
dispersed-minus-localized probability gap was closest to the eligible-set
median. This deterministic rule selects only the illustrative case; all
cohort-level analyses use the complete test set or stated stage subset.

\subsection{Repeated Localized Windows}

Table~\ref{tab:s-camelyon-k} reuses the same localized draws for prediction
averaging and exact patch union. The random control samples the same number of
patches as each slide's exact spatial union.

\begin{table*}[t]
\centering
\scriptsize
\setlength{\tabcolsep}{2.5pt}
\begin{tabular}{rrrrccc}
\toprule
$K$ & Mean patches & Coverage & \shortstack{Prediction\\average} &
\shortstack{Union before\\MIL} & \shortstack{Cardinality-matched\\random} \\
\midrule
1 & 128 & 3.45\% & $0.6202\pm0.0319$ / $0.5857\pm0.0056$ & $0.6202\pm0.0319$ / $0.5857\pm0.0056$ & $0.8831\pm0.0114$ / $0.8060\pm0.0089$ \\
2 & 250 & 6.60\% & $0.6971\pm0.0396$ / $0.6122\pm0.0414$ & $0.7161\pm0.0301$ / $0.6579\pm0.0051$ & $0.8865\pm0.0070$ / $0.8134\pm0.0211$ \\
5 & 601 & 15.62\% & $0.8139\pm0.0162$ / $0.5857\pm0.0056$ & $0.8282\pm0.0143$ / $0.7735\pm0.0133$ & $0.9133\pm0.0129$ / $0.8531\pm0.0256$ \\
10 & 1,105 & 27.45\% & $0.8327\pm0.0180$ / $0.5980\pm0.0091$ & $0.8443\pm0.0105$ / $0.8000\pm0.0056$ & $0.9660\pm0.0075$ / $0.8988\pm0.0244$ \\
20 & 1,900 & 44.41\% & $0.8968\pm0.0139$ / $0.6041\pm0.0046$ & $0.9116\pm0.0148$ / $0.8571\pm0.0250$ & $0.9701\pm0.0085$ / $0.9041\pm0.0406$ \\
\bottomrule
\end{tabular}
\caption{CAMELYON16 repeated-window response under five matched-model seeds.
Entries are AUROC / balanced accuracy; coverage is the fraction of each
reference bag represented by the exact union.}
\label{tab:s-camelyon-k}
\end{table*}

At $K=5$, union-minus-average is $+0.0143\pm0.0072$ AUROC and
$+0.1878\pm0.0155$ balanced accuracy, with all five seed differences positive.
The paired slide-bootstrap differences are $+0.0125$ (95\% CI $-0.0125$ to
$+0.0383$) and $+0.1939$ ($+0.1327$ to $+0.2653$), respectively. Random-minus-
spatial-union AUROC remains $+0.0851\pm0.0060$ at $K=5$ and
$+0.0585\pm0.0102$ at $K=20$, separating the aggregation-order gain from the
remaining geometry cost.

\subsection{Probability-Level Aggregation Diagnostics}

Table~\ref{tab:s-probability-quality} complements hard-decision QWK and balanced
accuracy with ranking and proper probability scores computed from the saved
per-case predictions. No model is retrained. On PANDA, union improves both
ordinal association and macro one-versus-rest ranking in every seed, while
lower log loss and ranked probability score show that the result is not only a
change in the argmax decision. On CAMELYON16, the large balanced-accuracy gain
is primarily an operating-point effect, although log loss also improves.

\begin{table*}[t]
\centering
\small
\setlength{\tabcolsep}{4.0pt}
\begin{tabular}{llrrrrr}
\toprule
Task ($K=5$) & Aggregation & Spearman $\uparrow$ & Macro AUROC $\uparrow$ & Log loss $\downarrow$ & RPS $\downarrow$ & Sensitivity $\uparrow$ \\
\midrule
PANDA & Prediction average & $0.9019\pm0.0008$ & $93.62\pm0.24$ & $0.810\pm0.003$ & $0.0617\pm0.0002$ & -- \\
PANDA & Union before MIL & $0.9331\pm0.0025$ & $95.45\pm0.26$ & $0.635\pm0.016$ & $0.0432\pm0.0014$ & -- \\
CAMELYON16 & Prediction average & -- & $81.39\pm1.62$ & $0.581\pm0.005$ & -- & $17.1\pm1.1$ \\
CAMELYON16 & Union before MIL & -- & $82.82\pm1.43$ & $0.482\pm0.020$ & -- & $54.7\pm2.7$ \\
\bottomrule
\end{tabular}
\caption{Probability-level aggregation diagnostics (model-seed mean $\pm$ SD).
PANDA macro AUROC is one-versus-rest and RPS is the normalized ranked
probability score. CAMELYON16 specificity is 100\% for both aggregation orders
at the fixed 0.5 threshold.}
\label{tab:s-probability-quality}
\end{table*}

\section{Human-Annotated Evidence Selection}

\subsection{Cohort Construction and Leakage Control}

Two ethics-approved deliveries organize images by assigned diagnosis and a
human binary judgment of whether the image contains visible evidence for that
diagnosis. We match image basenames to canonical case identifiers before any
split. Cases assigned to more than one diagnosis are excluded, as are classes
with fewer than 50 cases. The resulting H1/H2 cohorts contain 1,981/3,981 cases,
10/14 classes, and 3,447/4,715 annotated images. All 8,162 retained images map
to archived features; there are no missing cases or images. Among them, 202 H1
and 242 H2 cases contain both evidence-positive and evidence-negative images
and therefore support a within-case, equal-cardinality comparison.
The evidence labels are operational image-level judgments used to define the
test-time semantic selector; they are not training targets and are not treated
as consensus ground truth for diagnostic sufficiency.

Each hospital is modeled independently because its diagnosis vocabulary is
different; pooling hospitals would make site a label shortcut. A stratified
five-fold case split is fixed once. For each outer test fold, the next fold is
used for validation and the other three for training. No case crosses these
sets. Five model seeds are used. Training receives the complete annotated image
bag and its diagnosis. At test time, ten stable hash draws select one eligible
evidence-positive, evidence-negative, or random image from each mixed-evidence
case. The model and case are unchanged across these paired views.

Archived UNI v1 ViT-L/16 features were computed before this study. Each image
produces three deterministic 224-pixel center views: a native center crop, a
view after resizing the long edge to 448 pixels, and a 112-pixel center crop
resized to 224. Their 1,024-dimensional embeddings are concatenated to 3,072
dimensions. A 256-dimensional gated-attention classifier is trained for 30
epochs with AdamW ($10^{-4}$ learning rate and weight decay), inverse-frequency
class weights, and validation balanced-accuracy checkpoint selection. This
encoder differs from the UNI2-h WSI features in the public experiments and is
reported as a scope test rather than a directly pooled benchmark.

\subsection{Paired Results}

\begin{table*}[t]
\centering
\small
\setlength{\tabcolsep}{4.0pt}
\begin{tabular}{lrrrrrrrr}
\toprule
& & & \multicolumn{3}{c}{True-class probability} & \multicolumn{3}{c}{Balanced accuracy} \\
Hospital & All cases & Paired cases & Evidence+ & Random & Evidence$-$ & Evidence+ & Random & Evidence$-$ \\
\midrule
H1 & 1,981 & 202 & 46.92$\pm$1.49 & 38.28$\pm$1.10 & 30.94$\pm$0.70 & 59.09$\pm$1.18 & 48.77$\pm$1.73 & 40.63$\pm$2.52 \\
H2 & 3,981 & 242 & 53.53$\pm$0.99 & 42.07$\pm$0.51 & 30.81$\pm$0.55 & 71.95$\pm$2.29 & 57.65$\pm$1.36 & 41.90$\pm$2.02 \\
\bottomrule
\end{tabular}
\caption{Two-hospital human-evidence selection (\%, five-seed mean $\pm$ SD; ten
selector draws). Positive and negative views each contain one image from the
same held-out case.}
\label{tab:human-evidence-supp}
\end{table*}

Evidence-positive selection improves true-class probability and balanced
accuracy in all five seeds at both hospitals. Averaging model seeds and
selector draws before a 10,000-replicate case bootstrap stratified by hospital
and diagnosis gives a combined positive-minus-negative probability change of
$+19.65$ points (95\% CI $+16.90$ to $+22.43$). The corresponding
positive-minus-random change is $+10.17$ points ($+8.66$ to $+11.73$). These
labels independently show that semantic evidence availability changes the
paired model response; they do not
specify a WSI coordinate policy, measure pathologist ROI selection, or validate
the public-task mitigation controls.

\section{Three-Hospital Multi-Image Retrieval Audit}

\subsection{Ethics and Privacy}

The retrospective analysis was conducted under institutional ethics approval.
The committee name and approval identifier are omitted from this preprint
pending author confirmation and will be supplied in the final version. Hospitals are denoted
H1--H3, and raw reports, images, and case identifiers are not released. The
paper reports only aggregate metrics.

\subsection{Frozen Cohort and Leakage Control}

This separate audit freezes 5,000 cases with at least two images from each hospital. The
text target is the complete diagnostic report field embedded by the existing
Qwen3 feature extractor. It uses no masked report, regular-expression slot,
immunohistochemistry label, or manually inferred class. The extraction audit
finds 15,000 mapping entries and 15,000 feature groups.

Normalized-report hashes are assigned wholly to train, validation, or test
before hospital-specific retrieval galleries are formed. The split contains
8,500 training and 3,250 validation cases; the H1/H2/H3 test galleries contain
750/1,250/1,250 cases. No case or normalized-report hash crosses between
training and evaluation. Retrieval remains hospital-internal and does not test
cross-hospital alignment.

\begin{table}[t]
\centering
\small
\setlength{\tabcolsep}{3.2pt}
\begin{tabular}{lrrrr}
\toprule
Hospital & Cases & Median & Mean & Max. \\
\midrule
H1 & 5,000 & 2 & 2.088 & 27 \\
H2 & 5,000 & 2 & 2.010 & 6 \\
H3 & 5,000 & 2 & 2.258 & 29 \\
\bottomrule
\end{tabular}
\caption{Frozen private cohort.}
\label{tab:private-cohort}
\end{table}

\subsection{Image-Removal Response}

The primary metric is equivalence-aware image-to-text Recall@1: retrieval is
correct when the returned report has the same normalized report hash as the
query report. This avoids treating exact duplicate reports as false matches.
Matched image--report cosine is reported alongside it. Each entry is the mean
and sample standard deviation over model seeds 3407, 2027, and 17. Random
single-image and 50\%-retention policies average five deterministic selector
draws per seed.

\begin{table*}[t]
\centering
\small
\setlength{\tabcolsep}{4.2pt}
\begin{tabular}{llrrrr}
\toprule
Hospital & Policy & I2T R@1 & Drop & Matched cosine & Drop \\
\midrule
H1 & Full & 12.76$\pm$0.89 & -- & 50.22$\pm$1.10 & -- \\
   & Random single & 11.07$\pm$1.09 & 1.69 & 47.66$\pm$1.13 & 2.56 \\
   & Keep 50\% & 11.24$\pm$1.04 & 1.52 & 47.93$\pm$1.12 & 2.29 \\
   & Top-attention drop 50\% & 10.76$\pm$1.37 & 2.00 & 47.22$\pm$1.15 & 3.00 \\
\midrule
H2 & Full & 10.69$\pm$1.38 & -- & 54.98$\pm$1.08 & -- \\
   & Random single & 7.83$\pm$1.54 & 2.86 & 49.73$\pm$1.08 & 5.25 \\
   & Keep 50\% & 7.86$\pm$1.54 & 2.83 & 49.75$\pm$1.07 & 5.23 \\
   & Top-attention drop 50\% & 7.15$\pm$2.10 & 3.54 & 48.40$\pm$0.97 & 6.58 \\
\midrule
H3 & Full & 4.96$\pm$0.42 & -- & 51.48$\pm$0.88 & -- \\
   & Random single & 3.15$\pm$0.41 & 1.81 & 44.55$\pm$0.75 & 6.93 \\
   & Keep 50\% & 3.38$\pm$0.49 & 1.58 & 45.38$\pm$0.70 & 6.10 \\
   & Top-attention drop 50\% & 2.69$\pm$0.20 & 2.27 & 42.89$\pm$1.04 & 8.59 \\
\bottomrule
\end{tabular}
\caption{Attention MIL image-removal response (\%). Drops are relative to the
full multi-image case; larger positive drops indicate greater sensitivity.}
\label{tab:private-response}
\end{table*}

All nine hospital--policy comparisons degrade from full cases on both metrics.
H2 has the largest Recall@1 loss, whereas H3 has the largest matched-cosine
loss. These results support selection sensitivity in routine multi-image data,
but the report is not an independently adjudicated diagnostic label.

\subsection{Compatible Retrieval Objective}

For completeness, Table~\ref{tab:private-method} compares Attention MIL with
an earlier report-retrieval objective developed before the present measurement
framework. The latter is worse on H1,
better on H2, and approximately tied on H3. The private cohort therefore does
not provide a consistent method win and is used only as a phenomenon audit.

\begin{table*}[t]
\centering
\small
\setlength{\tabcolsep}{3.8pt}
\begin{tabular}{llrrrr}
\toprule
Hospital & Policy & Attention & Earlier retrieval objective & Difference & Favorable seeds \\
\midrule
H1 & Full & 12.76$\pm$0.89 & 12.09$\pm$1.08 & $-0.67$ & 0/3 \\
   & Random single & 11.07$\pm$1.09 & 10.82$\pm$0.91 & $-0.25$ & 2/3 \\
   & Keep 50\% & 11.24$\pm$1.04 & 10.92$\pm$0.86 & $-0.32$ & 2/3 \\
   & Top-attention drop 50\% & 10.76$\pm$1.37 & 10.36$\pm$0.28 & $-0.40$ & 2/3 \\
\midrule
H2 & Full & 10.69$\pm$1.38 & 11.97$\pm$0.41 & $+1.28$ & 3/3 \\
   & Random single & 7.83$\pm$1.54 & 8.25$\pm$0.43 & $+0.41$ & 2/3 \\
   & Keep 50\% & 7.86$\pm$1.54 & 8.26$\pm$0.42 & $+0.40$ & 2/3 \\
   & Top-attention drop 50\% & 7.15$\pm$2.10 & 7.28$\pm$0.97 & $+0.13$ & 2/3 \\
\midrule
H3 & Full & 4.96$\pm$0.42 & 4.96$\pm$0.37 & $+0.00$ & 1/3 \\
   & Random single & 3.15$\pm$0.41 & 3.19$\pm$0.19 & $+0.04$ & 2/3 \\
   & Keep 50\% & 3.38$\pm$0.49 & 3.52$\pm$0.27 & $+0.14$ & 2/3 \\
   & Top-attention drop 50\% & 2.69$\pm$0.20 & 2.85$\pm$0.32 & $+0.16$ & 2/3 \\
\bottomrule
\end{tabular}
\caption{Equivalence-aware image-to-text Recall@1 (\%). Positive differences
favor the earlier retrieval objective, which is distinct from the BagShift
measurement framework studied in the main paper.}
\label{tab:private-method}
\end{table*}

\section{Computing Environment and Metric Definitions}

The public-cohort experiments were run as single-GPU jobs on a server with two
AMD EPYC 9355 32-core processors, 384~GiB system memory, and NVIDIA RTX PRO
6000 Blackwell GPUs with 96~GiB memory, under Ubuntu 24.04. The production
environment used Python 3.12.3, PyTorch 2.13.0 with CUDA 13.0, NumPy 2.5.1,
pandas 3.0.3, scikit-learn 1.9.0, h5py 3.16.0, PyYAML 6.0.3, and einops 0.8.2.
Final configuration files specify the remaining model- and experiment-level
settings.

Quadratic weighted kappa (QWK) applies quadratic disagreement weights to the
ordered ISUP grades and is the primary PANDA metric because it distinguishes
nearby from distant grade errors. Balanced accuracy is the unweighted mean of
per-class recalls and therefore does not inherit the observed class
frequencies as weights. For lung subtyping, AUROC is computed from patient-level
probabilities and measures threshold-independent ranking; balanced accuracy
also reports behavior at the fixed decision threshold. Policy degradation is
defined by the paired metric-response equation in the main paper and is always
accompanied by the absolute reference and restricted-view levels.

\section{Release Boundary}

The accompanying code/data supplement contains exact public split manifests,
selector implementations, model configurations, aggregation and analysis code,
tests, and compact aggregate outputs. Raw public WSIs, large derived feature
tensors, pretrained weights, and checkpoints are excluded. Privacy restrictions
prevent release of the separate H1--H3 retrieval-audit images, reports,
features, and identifiers; its aggregate tables are the auditable outputs.
Aggregate H1/H2 results, evaluation code, split-generation logic, and the
annotation protocol will be released with the final artifact. Access to the
underlying images and binary annotations remains subject to institutional
governance and cannot be promised as an unconditional public release.